\documentclass[letterpaper, 10 pt, conference]{ieeeconf}  % Comment this line out if you need a4paper

\IEEEoverridecommandlockouts                              % This command is only needed if 
\usepackage{cite}
\usepackage{ragged2e}
\usepackage{amsmath,amssymb,amsfonts}
\usepackage[ruled,linesnumbered]{algorithm2e}
\usepackage{algorithmic}
\usepackage{graphicx}
\usepackage{textcomp}
\usepackage{multirow}
\usepackage[most]{tcolorbox} % powerful colored boxes
\usepackage{xcolor}
\usepackage[font=footnotesize, labelfont=footnotesize]{caption}
\usepackage{subcaption}

\usepackage{amsmath, xcolor, subcaption}
\definecolor{c1}{RGB}{192,0,0}     % C1 red
\definecolor{c2}{RGB}{237,125,49}  % C2 orange
\definecolor{c3}{RGB}{0,153,0}     % C3 green
\definecolor{c4}{RGB}{68,114,196}  % C4 blue
\definecolor{c5}{RGB}{128,128,128} % C5 gray

\newcommand{\pos}[1]{\textcolor{blue}{#1}}
\newcommand{\nega}[1]{\textcolor{red}{#1}}

\def\BibTeX{{\rm B\kern-.05em{\sc i\kern-.025em b}\kern-.08em
    T\kern-.1667em\lower.7ex\hbox{E}\kern-.125emX}}

\title{\LARGE \bf
Converting Sequenced Fuzzy Cognitive Maps to Causal Virtual Worlds with Large Video Generators
}

\author{Akash Kumar Panda$^{1}$ and Olaoluwa Adigun$^{2}$ and Bart Kosko$^{3}$% <-this % stops a space
\thanks{$^{1}$Akash Kumar Panda is with the Deparment of Electrical and Computer Engineering,
        University of Southern California, Los Angeles, CA, USA
        {\tt\small akashpan@usc.edu}}%
\thanks{$^{2}$Olaoluwa Adigun is with the School of Computing and Information Sciences, Florida International University
        Miami, FL 33199, USA
        {\tt\small olaadigu@fiu.edu}}%
\thanks{$^{3}$Bart Kosko is with the Deparment of Electrical and Computer Engineering,
        University of Southern California, Los Angeles, CA, USA
        {\tt\small kosko@usc.edu}}%
}

\begin{document}
\newtcolorbox{llmbox}[1][]{
  enhanced,
  breakable,                 % allow page breaks inside
  arc=4pt, %sharp corners,
  colback=blue!3,            % background
  colframe=blue!10!black,    % border color
  coltitle=blue!20!black,
  coltext=red!50!black, 
  %fontupper=\ttfamily\small, % monospaced content
  fontupper=\small\justifying, % <-- fully justified text
  fonttitle=\bfseries\ttfamily,
  left=1.2ex, right=1.2ex, top=0.8ex, bottom=1ex,
  boxsep=0.6ex,
  %borderline west={0pt}{0pt}{blue!60!black}, % accent bar
  title=LLM Output,          % default title
  #1                         % allow user overrides
}

\newtcolorbox{human_annotation}[1][]{
  enhanced,
  breakable,                 % allow page breaks inside
  arc=4pt, %sharp corners,
  colback=blue!3,            % background
  colframe=blue!60!black,    % border color
  coltitle=blue!20!black,
  coltext=red!50!black, 
  %fontupper=\ttfamily\small, % monospaced content
  fontupper=\small\justifying, % <-- fully justified text
  fonttitle=\bfseries\ttfamily,
  left=1.2ex, right=1.2ex, top=0.8ex, bottom=1ex,
  boxsep=0.6ex,
  %borderline west={0pt}{0pt}{blue!60!black}, % accent bar
  title=Example Annotation Output,          % default title
  #1                         % allow user overrides
}

\maketitle
\thispagestyle{empty}
\pagestyle{empty}

%%%%%%%%%%%%%%%%%%%%%%%%%%%%%%%%%%%%%%%%%%%%%%%%%%%%%%%%%%%%%%%%%%%%%%%%%%%%%%%%
\begin{abstract}

We show how users can create and manipulate causal virtual worlds with large-language-model (LLM) and large-video-model agents.
The approach uses feedback fuzzy cognitive maps (FCMs) both to model the granular causal structure of the virtual world and to guide its causal evolution.
The local causal rules are partial or fuzzy while the FCM's feedback structure produces global equilibria that define causal scenarios.
A sequence of \emph{dynamical} meta-rules of the form ``If $\mathcal{A}$ then $\mathcal{B}$" define the causal scenes of the virtual-world video.
The if-part causal pattern $\mathcal{A}$ perturbs the FCM's virtual world at the user's or agent's discretion.
The FCM's transient feedback dynamics define the meta-rule's causal arrow of implication.
The then-part $\mathcal{B}$ is the resulting equilibrium attractor such as a FCM limit cycle or fixed point. 
Our algorithm extracts these meta-rules from the FCM and guides the LLM agent to write a script based on the FCM meta-rule sequence. 
The large video generator converts the meta-rule into a video scene in accord with the flow of the dynamics.
We applied the agent-based technique to a simple FCM that describes an undersea world of dolphins and sharks.
Google's Gemini 3.1 generated the script and Google's Veo 3.1 generated the dolphin-shark video.
The approach is general and can scale by mixing larger FCMs and AI agents to produce more immersive virtual worlds.

\end{abstract}

%%%%%%%%%%%%%%%%%%%%%%%%%%%%%%%%%%%%%%%%%%%%%%%%%%%%%%%%%%%%%%%%%%%%%%%%%%%%%%%%
\section{Agentic AI Generation of Virtual Worlds from Sequenced Fuzzy Cognitive Maps}\label{Introduction}

We show how to generate realistic causal virtual-world videos by combining feedback Fuzzy Cognitive Maps (FCMs) with large-video-generator agents.

This new virtual-world technique is general and scales for arbitrary mixed causal-graph FCMs combined with controlled AI agents.
As OpenAI has stated:  ``scaling video generation models is a promising path towards building general purpose simulators of the physical world."\cite{videoworldsimulators2024}
Sequenced FCM dynamics offers a practical way to approximate such models.

\begin{figure*}[ht]
\centering
\includegraphics[width=0.95\linewidth]{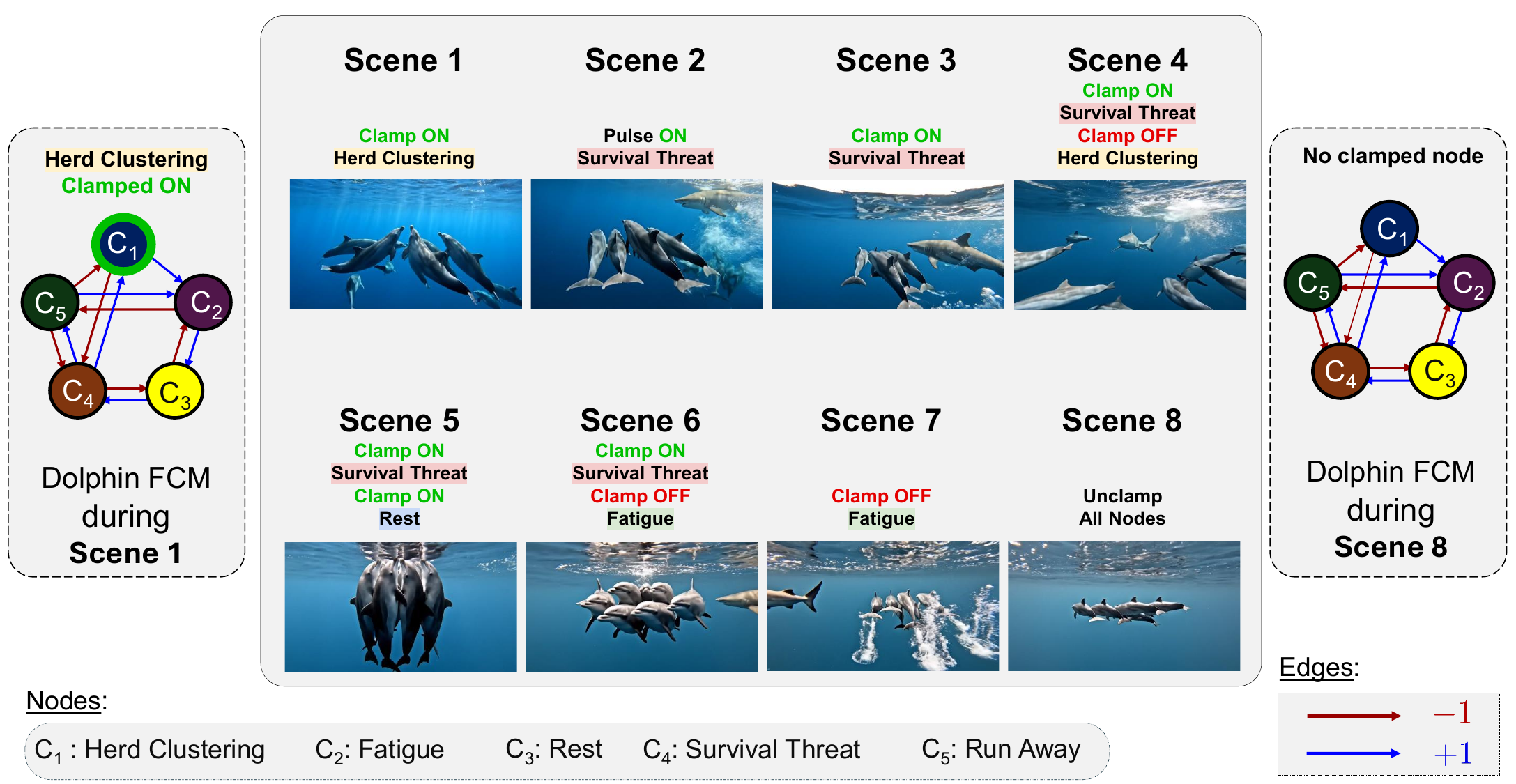}
\caption{Frames from a video of the virtual world generated from a 5-node Fuzzy Cognitive Map (FCM) that models dolphin behavior in presence of a ``Survival Threat''. 
The video shows a pod of dolphins running away from a shark over 8 scenes.
The input stimulus for each scene is above the frame corresponding to the scene. 
The figures on the left and right of the video frames show the respective state of the FCM at the beginning and the end of the video. 
The figure highlights clamped node ``Herd Clustering'' in green.}
\label{fig:figure1}
\end{figure*}

Figure~\ref{fig:figure1} shows the system flow of stimulated FCM causal patterns that produce an AI-generated 1-minute video from a simple 5-node FCM.
The FCM defines a coarse-grained undersea world of dolphins and sharks.
We use this simple dolphin FCM because over the decades it has become a type of small-scale test and teaching model of an FCM \cite{dickerson1994virtual}.
The FCM itself is a nonlinear feedback dynamical system whose causal graph is signed and weighted or partial to reflect partial or fuzzy causality \cite{kosko1986fuzzy,kosko1988hidden,osoba2017fuzzy,ziv2018potential,glykas2010fuzzy,papageorgiou2013fuzzy,stach2010divide,kosko1988hidden,taber2007quantization}.
The local causal rule structure of the FCM helps make the nonlinear system interpretable or explainable \cite{8466590, samek2019explainable}.  

\begin{figure*}[htbp]
\begin{subfigure}{0.42\linewidth}
\centering
\includegraphics[height=0.5\textwidth, width=1.0\textwidth]{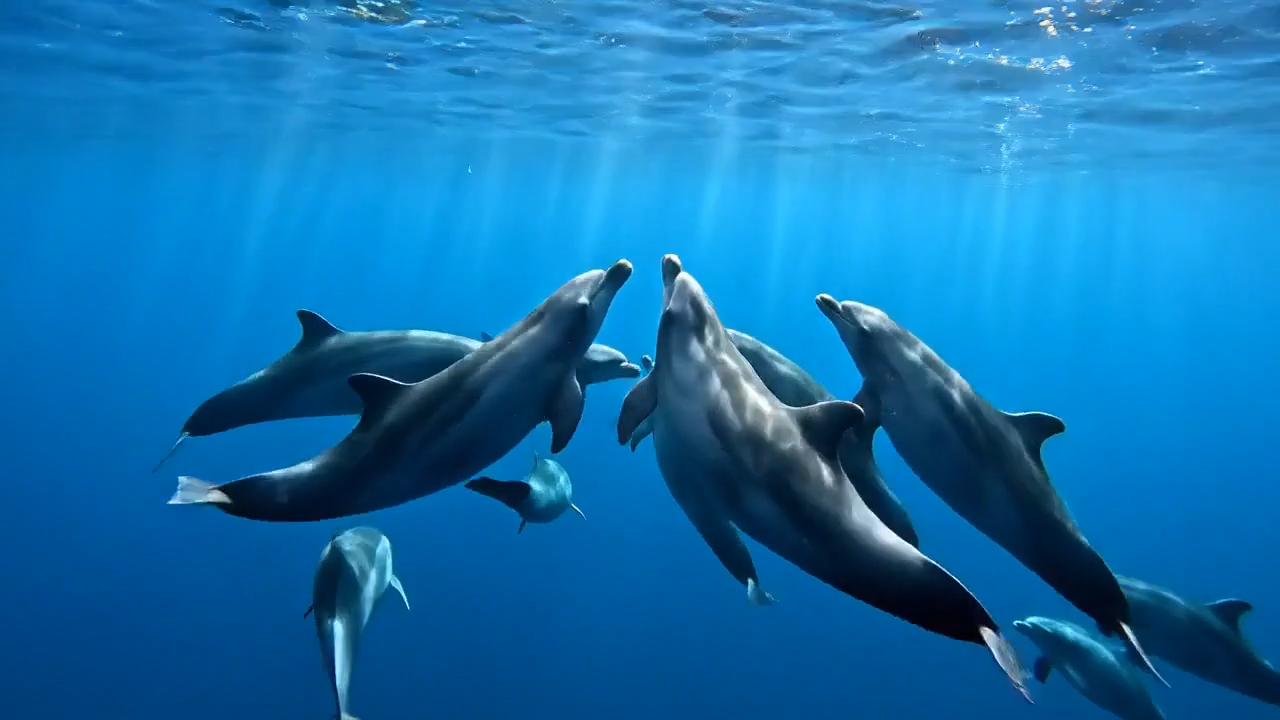}
\caption{{Scene 1}}
\label{fig:shark_dolphin_scene_1}
\vspace{0.1in}
\end{subfigure}
%\hfill
\begin{subfigure}{0.42\linewidth}
\centering
\includegraphics[height=0.5\textwidth, width=1.0\textwidth]{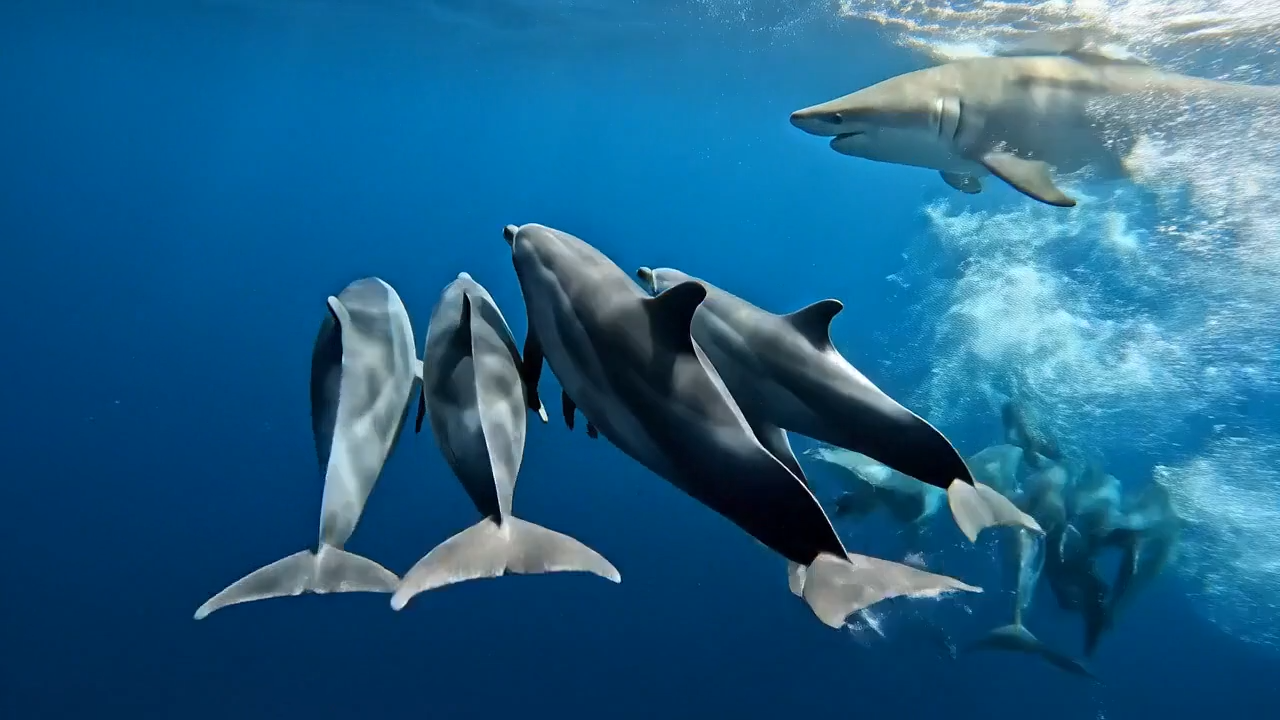}
\caption{{Scene 2}}
\label{fig:shark_dolphin_scene_2}
\vspace{0.1in}
\end{subfigure}
\hfill
\begin{subfigure}{0.42\linewidth}
\centering
\includegraphics[height=0.5\textwidth, width=1.0\textwidth]{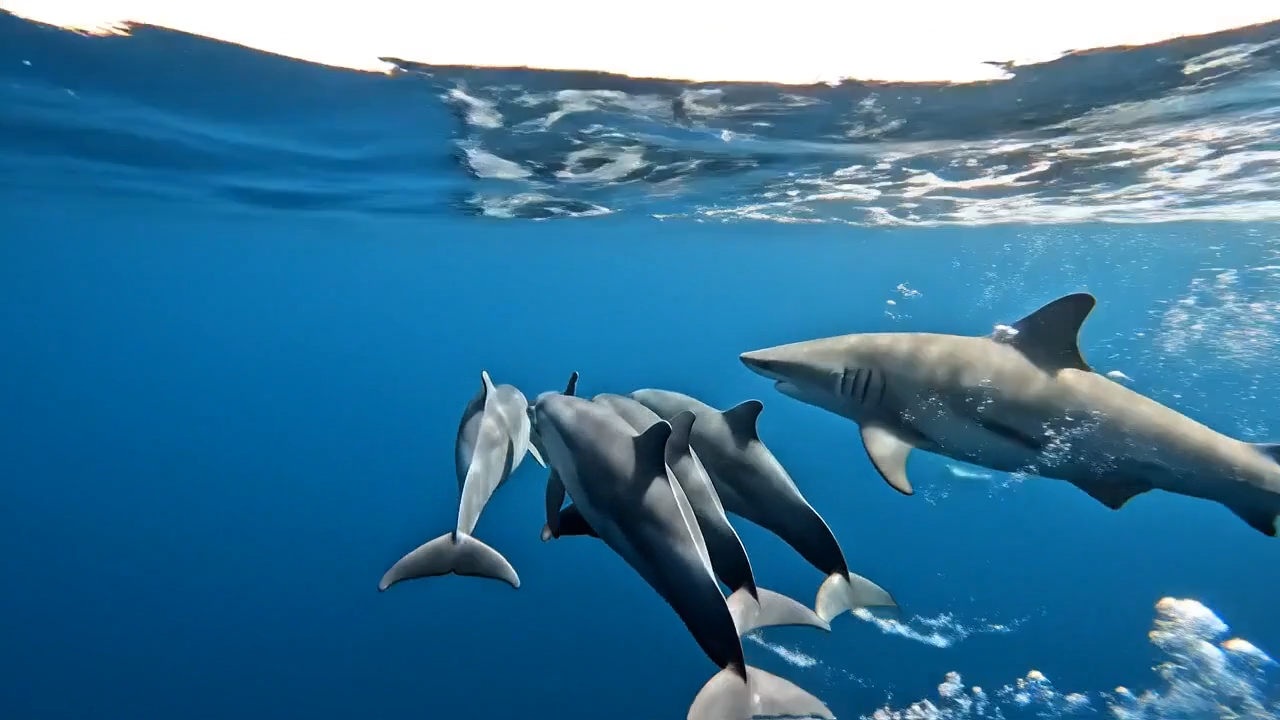}
\caption{{Scene 3} }
\label{fig:shark_dolphin_scene_3}
\vspace{0.1in}
\end{subfigure}
\hfill
\begin{subfigure}{0.42\linewidth}
\centering
\includegraphics[height=0.5\textwidth, width=1.0\textwidth]{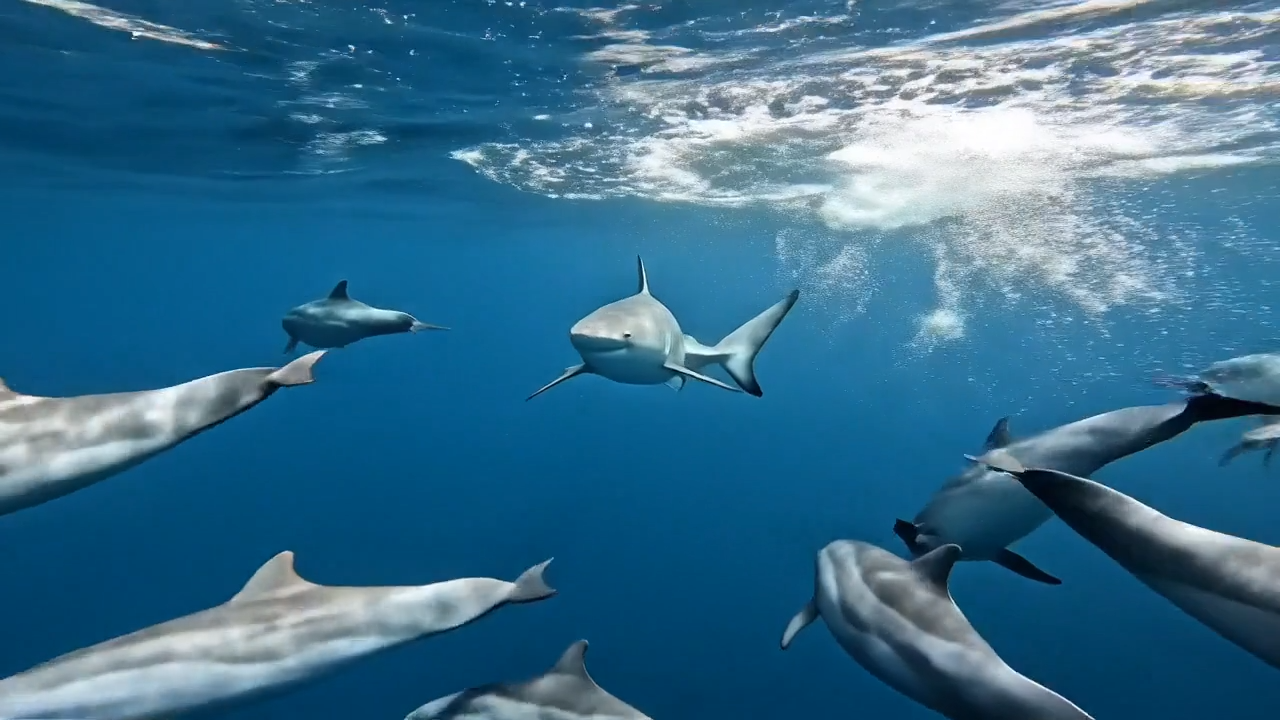}
\caption{{Scene 4} }
\label{fig:shark_dolphin_scene_4}
\vspace{0.1in}
\end{subfigure}
\begin{subfigure}{0.42\linewidth}
\includegraphics[height=0.5\textwidth, width=1.0\textwidth]{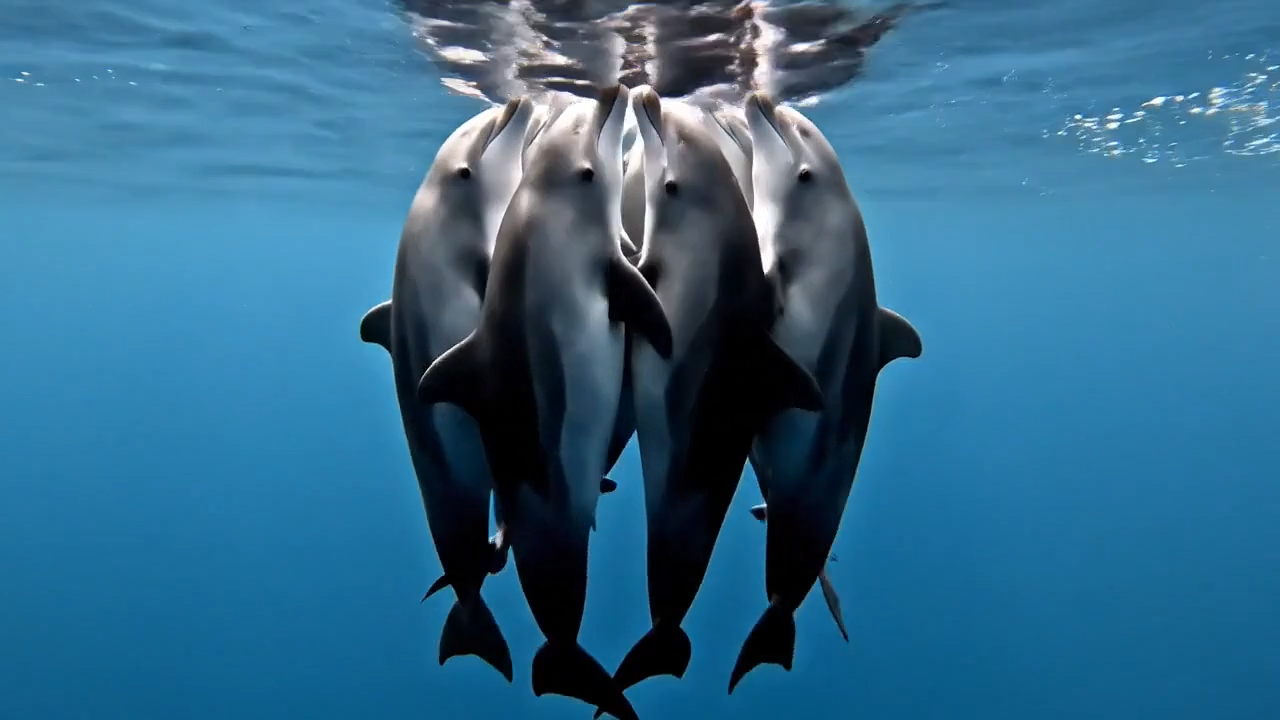}
\caption{{Scene 5}}
\label{fig:shark_dolphin_scene_5}
\vspace{0.1in}
\end{subfigure}
\hfill
\begin{subfigure}{0.42\linewidth}
\centering
\includegraphics[height=0.5\textwidth, width=1.0\textwidth]{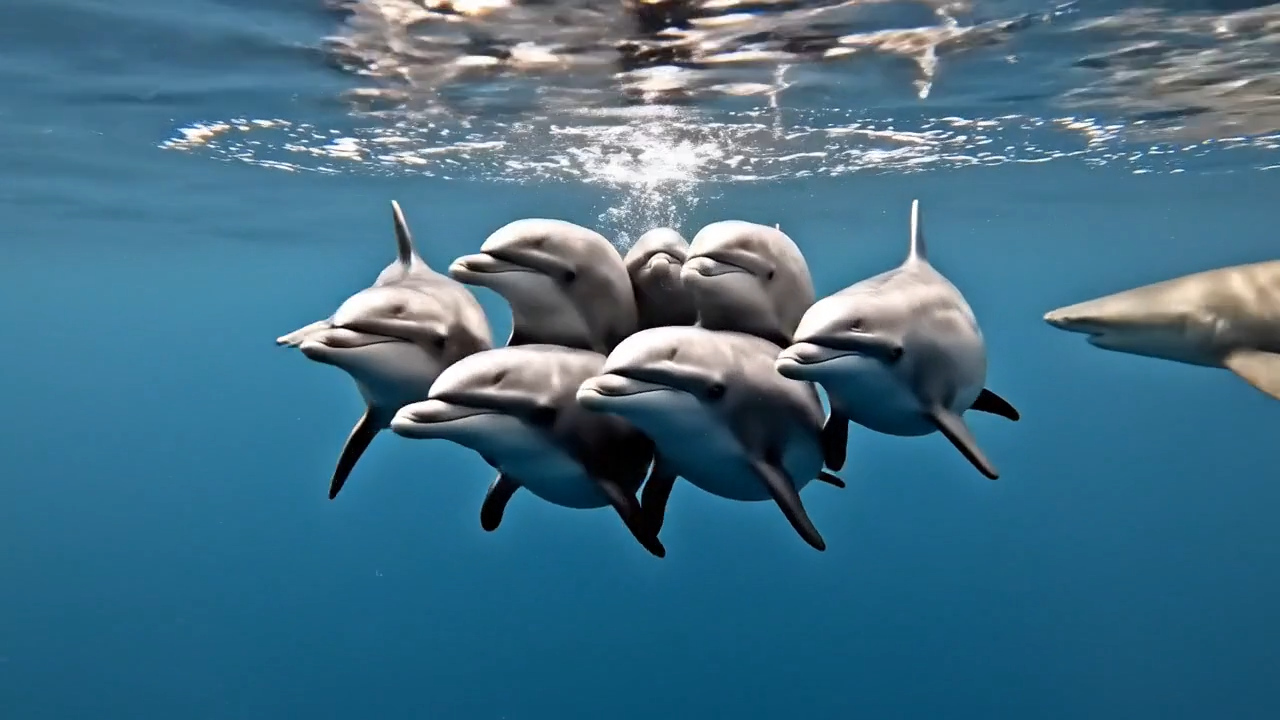}
\caption{{Scene 6}}
\label{fig:shark_dolphin_scene_6}
\vspace{0.1in}
\end{subfigure}
\hfill
\begin{subfigure}{0.42\linewidth}
\centering
\includegraphics[height=0.5\textwidth, width=1.0\textwidth]{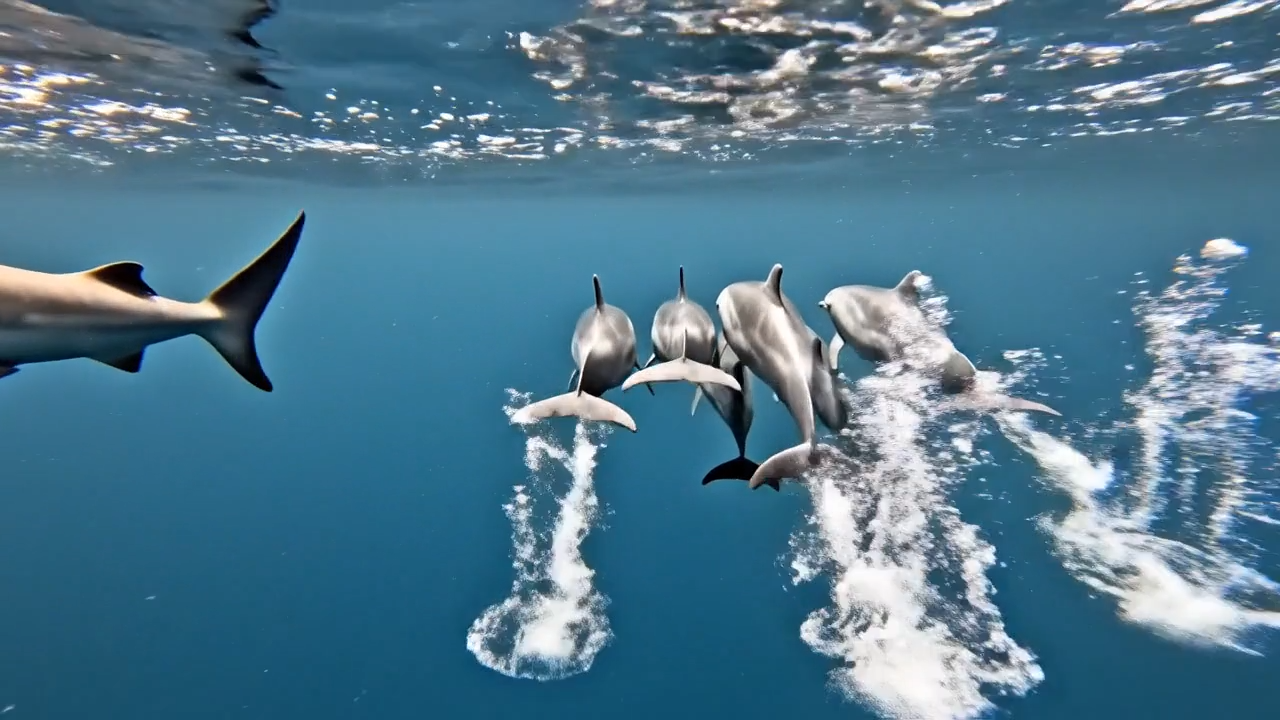}
\caption{{Scene 7}}
\label{fig:shark_dolphin_scene_7}
\vspace{0.1in}
\end{subfigure}
\hfill
\begin{subfigure}{0.42\linewidth}
\centering
\includegraphics[height=0.5\textwidth, width=1.0\textwidth]{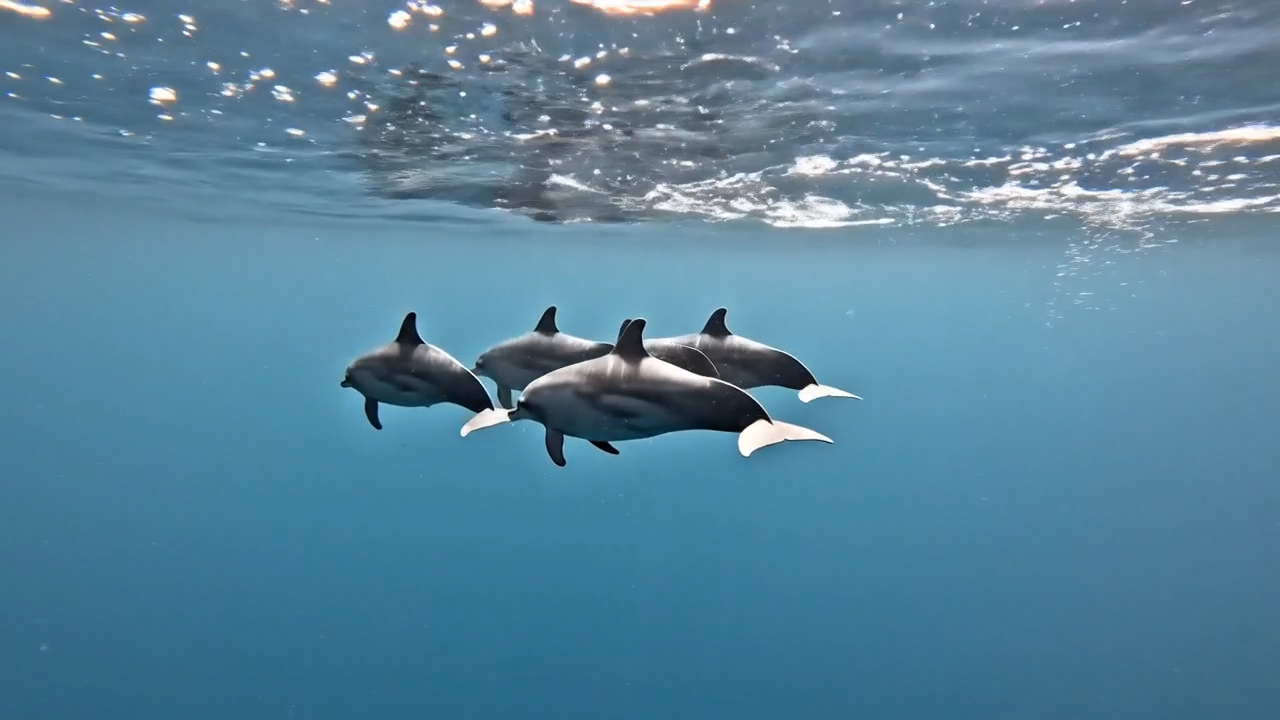}
\caption{{Scene 8}}
\label{fig:shark_dolphin_scene_8}
\vspace{0.1in}
\end{subfigure}
\caption{Limit cycle 1: Interaction of a dolphin pod and a tiger shark in a water. 
(a). Scene 1: The dolphin pod rests and plays.
(b). Scene 2: The dolphin pod runs away from a threat, gets tired, and rests.
(c). Scene 3: The dolphins swim in a herd cluster to avoid a large tiger shark.
(d). Scene 4: The tiger shark breaks the dolphins' herd-cluster apart as the dolphins tire.
(e). Scene 5: The tired dolphins stop running away and regroup in a tighter herd-cluster.
(f). Scene 6: The dolphins push through their fatigue and swim away from the large tiger shark. 
(g). Scene 7: The dolphin pod gets away from the large tiger shark.
(h). Scene 8: The dolphin goes back to playing and resting until their next shark encounter.}
\label{fig:Figure2}
\end{figure*}

The FCM-generated video shows how a pod of playful dolphins responds to the threat of a large tiger shark.
The video uses a sequence of 8 scenes as we explain in detail in the next two sections.
Figure~\ref{fig:Figure2} shows further snapshots from the video.

Figure~\ref{fig:DolphinFCM} shows the 5-node dolphin-shark FCM and its corresponding 5-by-5 causal edge matrix $E$.  
This paper uses just this small FCM to illustrate the process of generating virtual worlds although we stress again that the process is completely general for any size or mixture of FCMs.

Figure~\ref{fig:figure4} shows the sequence of 8 scenes as 8 input-output pairs of stimulated FCM states that map to equilibrium limit cycles or fixed-point attractors. 
This gives a type of dynamical system ``storyboard" of the FCM-based video or virtual world.
Users can pick the stimulus patterns directly or ask the AI agent to pick the nearest-matching patterns for them.
Users can also allow an agent to pick the stimulus patterns itself.

Figures 5-8 show frames from the first 4 scenes of the final video.
Figure 9 shows all sequenced limit cycles used.

The key insight is that a FCM dynamical system defines a set of causal \emph{meta-rules} of the form $\mathcal{A}_j \rightarrow \mathcal{B}_j$.
We ``sequence" or impose a time order on these FCM meta-rules.
We then use the sequenced meta-rules to drive the time-ordered scenes of the generated video or virtual world.

The nonlinear FCM dynamical system $\mathcal{F}$ itself is a time-varying mapping $\mathcal{F}:  \mathbb{R}^n \rightarrow \mathbb{R}^n$ from a causal pattern space such as $\mathbb{R}^n$ back into itself.
A pulsed or sustained (``clamped") input vector $x \in \mathbb{R}^n$ produces a sequence of transient causal states that ends in an equilibrium attractor that can range from a fixed point to a limit cycle to aperiodic chaos.
The simple binary-state dolphin FCM in Figure 3 always ends in a limit cycle or a fixed-point attractor that lives inside the binary 5-cube $\{0, 1\}^5$.
The fuzzy causal edge values $e_{ij}$ of the directed causal edge from concept node $C_i$ to node $C_j$ takes values in the bipolar interval $[-1, 1]$ in general where positive values $e_{ij} > 0$ denote causal increase and negative values $e_{ij} < 0$ denote causal decrease and $e_{ij} = 0$ denotes no causal effect.

The global meta-rule $\mathcal{A}_j \rightarrow \mathcal{B}_j$ asks and answers its own ``what-if" question:  What dynamical equilibrium $\mathcal{B}_j$ results if we perturb the FCM causal system with new input $x$ or otherwise knock it into a new state of \emph{disequilibrium} $\mathcal{A}_j$?

So the implication arrow ``$\rightarrow$" in the meta-rule $\mathcal{A}_j \rightarrow \mathcal{B}_j$ is not a classical logical or material-implication operator.
The arrow is instead a \emph{causal implicant} that depends on system transients and hence depends on time in general.
These are unconstrained rules.
The user can add \emph{constraint} rules that guide or extend the equilibria and so modify the FCM video.

The resulting set of such dynamical meta-rules $\mathcal{A}_j \rightarrow \mathcal{B}_j$ endows the FCM system a with a global level of explainability or XAI.
This still holds in the general case when we take convex combinations $w_1 \mathcal{F}_1 + \cdots  + w_K \mathcal{F}_K$ of $K$-many FCMs $\mathcal{F}_1, \ldots , \mathcal{F}_K$ for nonnegative mixing weights $w_j$ that sum to unity and that can depend on $x$.
 Mixing FCM causal edge matrices $E_1, \ldots , E_K$ always produces a new causal edge matrix and thus always produces an FCM $\mathcal{F}$.
Appropriate zero-padding of rows and columns ensures that the mixed causal edge matrices are conformable for matrix addition.

This mixing-closure property greatly extends the application reach and knowlege-representation power of FCMs.
Users can combine and uncombine and adapt the underlying mixed FCM at will.
Users can also pick the mixed meta rules and especially the disequilibrium stimuli $\mathcal{A}_j$ at will to create and change the virtual world.
Agentic AI systems can further assist in this immersive world-creating process.

The next section reviews FCMs and show how to sequence and fire their meta-rules.
It shows how the dolphin FCM's 8 causal meta-rules $\mathcal{A}_1 \rightarrow \mathcal{B}_1, \ldots , \mathcal{A}_8 \rightarrow \mathcal{B}_8$ become the 8 scene prompts to the AI video generator. 

The last section explains how to convert these FCM scenes to prompts that control Google's Veo 3.1 large-video generator by way of Algorithm 1 below. 
We here focus on user-driven choices of meta-rules $\mathcal{A}_j \rightarrow \mathcal{B}_j$ and resulting scenes but recent work shows that we can easily use large language models (LLMs) to generate FCM knowledge graphs from text or vice versa \cite{panda2025causal} in more complex agentic AI systems.

\section{Sequenced Fuzzy Cognitive Maps}

\subsection{Fuzzy Cognitive Maps (FCM)}

FCMs model causal dynamical systems as directed weighted graphs\cite{kosko1986fuzzy,kosko1988hidden,osoba2017fuzzy,ziv2018potential,glykas2010fuzzy,papageorgiou2013fuzzy,stach2010divide,kosko1988hidden,taber2007quantization}. 
The concept nodes describe the causal variables in the dynamical system and the directed edges describe the causal relationships between those nodes. 
FCMs allow feedback and therefore converge to non-trivial equilibria like limit cycles. 
FCMs model dynamical systems by approximating their underlying maps from inputs to equilibria. 

\subsection{Causal Edge Matrix}

The directed edges of the FCM describe the causal relationships between concept nodes. 
An edge $e_{ij}$ from the $i^{\text{th}}$ concept node $C_i$ to the $j^{\text{th}}$ concept node $C_j$ means ``$C_i$ causes $C_j$''. 
The fuzzy edge weights describe partial causality and we use the same symbol for this value---it can also vary with time.
The causal edge weight $e_{ij} \in [-1,1]$ on the $ij$th edge states the degree to which $C_i$ causes $C_j$:
\begin{align}
    e_{ij} = Degree(C_i \rightarrow C_j).
\end{align}
A positive $e_{ij}$ means that $C_j$ increases when $C_i$ increases and a negative $e_{ij}$ means that $C_j$ decreases if $C_i$ increases. 
The weight $e_{ij}$ is zero when there is no causal edge between $C_i$ and $C_j$. 
The magnitude of $e_{ij}$ is high if there is a strong causal relationship between $C_i$ and $C_j$. 
A low magnitude of $e_{ij}$ describes a weak causal relationship between $C_i$ and $C_j$.

An $n \times n$ matrix $E$ describes all the directed weighted edges of a $n$-node FCM. 
The causal edge weight $e_{ij}$ corresponds to the matrix element on the $i^{\text{th}}$ row and the $j^{\text{th}}$ column. 
The matrix element is zero if there is no edge between the corresponding node-pair. 

\begin{figure}[ht]
\centering
%\begin{subfigure}[b]{0.5\textwidth}
%\includegraphics[scale=0.25]{images/Dolphin-FCM-Graph.png}
%\end{subfigure}
\begin{subfigure}[b]{0.5\textwidth}
\includegraphics[scale=0.3]{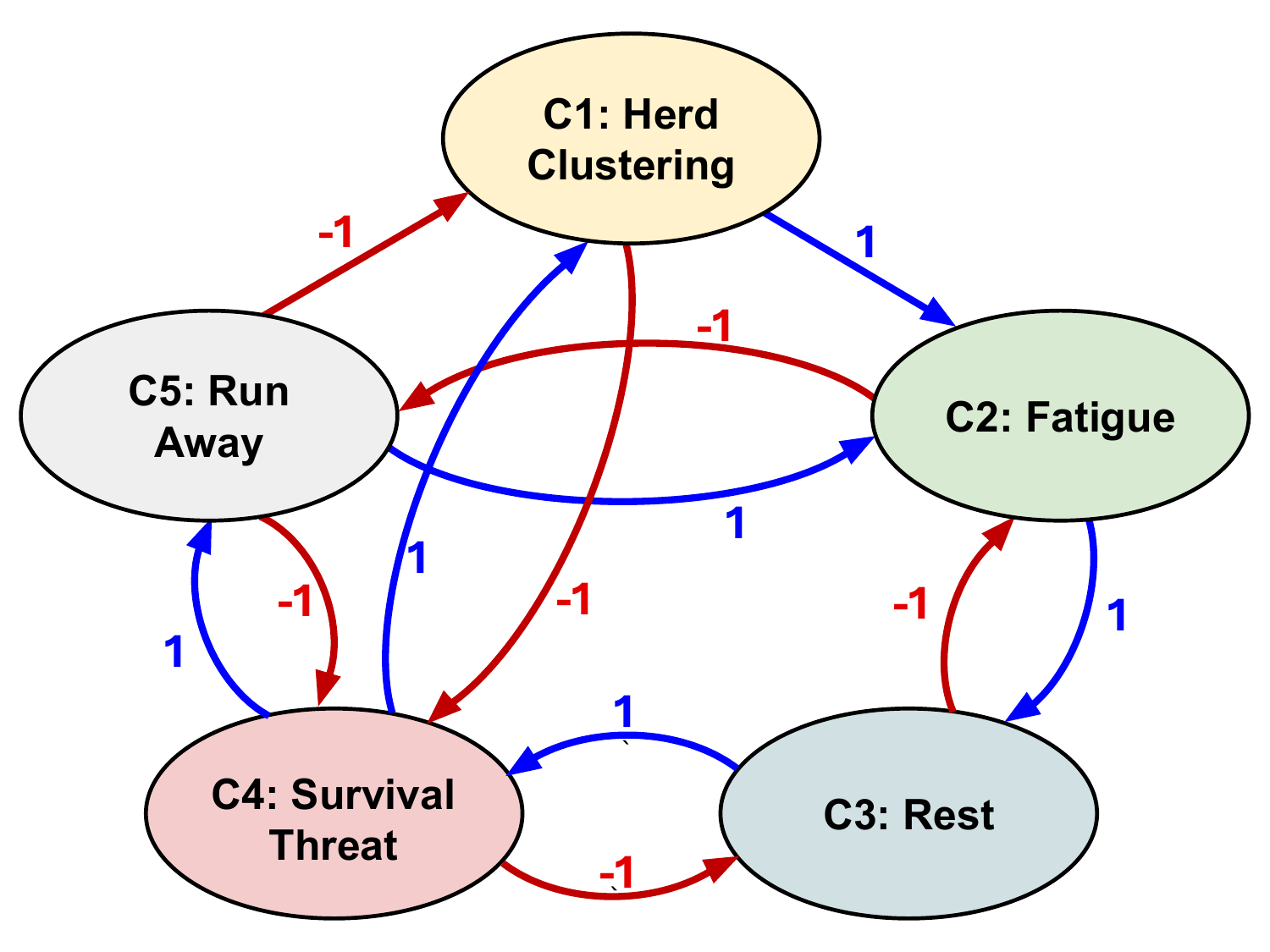}
 \caption{ }
\end{subfigure}

\begin{subfigure}[b]{0.5\textwidth}
\vspace{0.2in}
  \centering
  $E =
  \begin{array}{c@{\hspace{0pt}}c}
    & \begin{array}{ccccc}
        \textcolor{c1}{\mathrm{C_1}} & \textcolor{c2}{\mathrm{C_2}} &
        \textcolor{c3}{\mathrm{C_3}} & \textcolor{c4}{\mathrm{C_4}} &
        \textcolor{c5}{\mathrm{C_5}}
      \end{array} \\[2pt]
    \begin{array}{c}
      \textcolor{c1}{\mathrm{C_1}} \\ \textcolor{c2}{\mathrm{C_2}} \\
      \textcolor{c3}{\mathrm{C_3}} \\ \textcolor{c4}{\mathrm{C_4}} \\
      \textcolor{c5}{\mathrm{C_5}}
    \end{array} &
    \begin{bmatrix}
      0        & \pos{1}  & 0        & \nega{-1} & 0        \\
      0        & 0        & \pos{1}  & 0        & \nega{-1} \\
      0        & \nega{-1} & 0        & \pos{1}  & \nega{-1} \\
      \pos{1}  & 0        & \nega{-1} & 0        & \pos{1}  \\
      \nega{-1} & \pos{1}  & 0        & \nega{-1} & 0
    \end{bmatrix}
  \end{array}$
  \caption{ }
  \label{fig:edge-matrix}
\end{subfigure}
\caption{The 5-node Dolphin FCM with trivalent causal edges. 
(a) The directed weighted graph for the dolphin FCM. 
The directed edge $e_{ij}$ is trivalent because $e_{ij} \in \{-1, 0, 1\}$ but in general can be properly fuzzy and take on any directed causal degree in the bipolar interval $[-1, 1]$.
The positive edges are in blue and the negative edges are in red. 
(b) The causal edge matrix $E$ that corresponds to the dolphin FCM. 
The 5 concept nodes $C_1$-$C_5$ index both the rows and the columns. 
The matrix elements give the edge weights $e_{ij}$. 
The positive edge values $e_{ij} > 0$ are in blue and the negative edge values $e_{ij} < 0$ are in red. }
\label{fig:DolphinFCM}
\end{figure}

Figure~\ref{fig:DolphinFCM} shows the dolphin FCM with 5 nodes: ``Herd Clustering'', ``Fatigue'', ``Rest'',  ``Survival Threat'', and ``Run Away''. 
The figure also shows its corresponding causal edge matrix. 
The matrix element $e_{45}$ on the $4^{\text{th}}$ row and the $5^{\text{th}}$ column is 1 because there is a positive edge with weight 1 from the $4^{\text{th}}$ concept node ``Survival Threat'' to the $5^{\text{th}}$ concept node ``Run Away''. 
This says that the dolphins ``run away'' when a ``survival threat'' like a shark is present. 

\begin{figure*}[ht]
\centering
\includegraphics[scale=0.27]{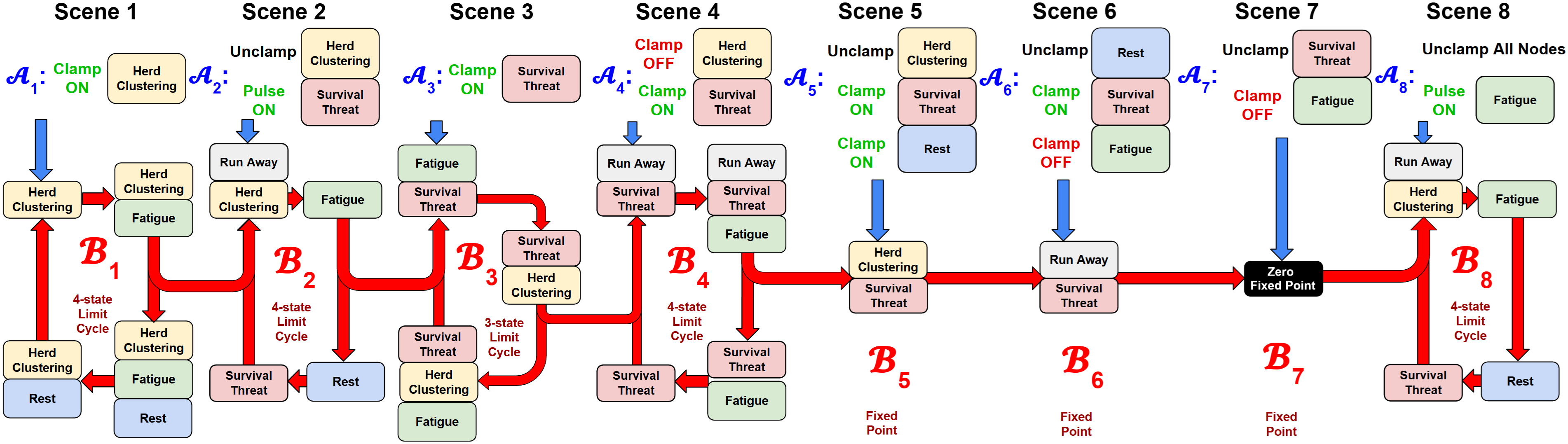}
\caption{Sequenced dolphin FCM Equilibria:  A sequence of 8 input patterns and clamping conditions $\mathcal{A}_k$ stimulate the dolphin FCM into their respective equilibria $\mathcal{B}_k$. 
The scene $k$ stimulates the FCM with$\mathcal{A}_k$. 
This breaks the FCM's prior equilibrium $\mathcal{B}_{k-1}$ and pushes it into its new equilibrium $\mathcal{B}_k$. }
\label{fig:figure4}
\end{figure*}

\subsection{FCM Evolution}

A $n$-dimensional row vector $C(t) \in [0,1]^n$ describes the state of the FCM's concept nodes at time $t$. 
The $i^{\text{th}}$ node is ``active'' or ``on'' at time $t$ if the $i^{\text{th}}$ component $C_i(t)$ of the state vector $C(t)$ is equal to or close to one. 
The $i^{\text{th}}$ node is ``inactive'' or ``off'' at time $t$ if $C_i(t)$ is equal to or close to zero. 
A node is partially active otherwise. 
The causal variables corresponding to the active nodes are present in the system and those corresponding to the inactive nodes are absent. 
The causal factors are partially present in the system if their corresponding nodes are partially active. 

Consider the dolphin FCM in figure~\ref{fig:DolphinFCM}. 
If $C(t) = \begin{pmatrix} 0 & 0 & 0 & 1 & 0 \end{pmatrix}$ then the $4^{\text{th}}$ node ``Survival Threat'' is ``on'' and all other nodes are ``off'' at time $t$. 
This means there is a ``threat'' like a shark lurking around. 

FCMs evolve in discrete time through vector-matrix multiplication and nonlinear compression--at least in the simple FCM we use for this video creation. 
The state $C_j(t+1)$ of the $j^{\text{th}}$ concept node $C_j$ at discrete time step $t+1$ is:
\begin{align}\label{eq:fcm-update}
    C_j(t+1) = \Phi\bigg(\sum_{i=1}^n  C_i(t)  \; e_{ij}\bigg)
\end{align}
if $\Phi$ is a nonlinear function bounded between zero and one. 

The sum $\sum_{i=1}^n  C_i(t) \; e_{ij}$ is the matrix product between the state row-vector $C(t)$ and the edge matrix $E$. 
The nonlinear function $\Phi$ then compresses this product between zero and one. 
We stress that $\Phi$ is an arbitrary but bounded nonlinear function in general even though here we use the simplest case of a threshold nonlinearity.

This process repeats itself to give the discrete-time evolution of the FCM. 
The FCM starts with the initial state $C(0)$ at time $t = 0$ and then goes through the states $C(1)$, $C(2)$, $C(3)$, and so on in order. 
The active nodes in this state-vector sequence qualitatively describe the trajectory of the dynamical system that the FCM models. 

Consider the dolphin FCM with an initial state $C(0) = \begin{pmatrix} 0 & 0 & 0 & 1 & 0 \end{pmatrix}$. 
Equation~\ref{eq:fcm-update} and figure~\ref{fig:DolphinFCM}(b) give $C(1) = \begin{pmatrix} 1 & 0 & 0 & 0 & 1 \end{pmatrix}$. 
``Survival Threat'' was active at $t=0$ and ``Herd Clustering'' and ``Run Away'' were active at $t=1$.  
This shows the dolphins' behavior in the presence of a shark. 
They group themselves together in herd clusters and run away from the shark.

\subsection{FCM Equilibria}

The equilibria characterize a dynamical system. 
The equilibrium behavior of the FCM depends on the limiting behavior of the state-vector sequence. 
The FCM converges to a ``fixed point'' if the state-vector sequence converges to a constant vector. 
The FCM converges to a $K$-step ``limit cycle'' for an integer $K > 1$ if $C(t+K) = C(t)$ somewhere in the state-vector sequence. 
Then the FCM converges to an equilibrium where $K$ state vectors repeat themselves over and over in the same order. 
The FCM may also converge to a chaotic attractor where there are no repeating patterns in the state-vector sequence. 

Consider the dolphin FCM with same initial state $C(0) = \begin{pmatrix} 0 & 0 & 0 & 1 & 0 \end{pmatrix}$. 
Equation~\ref{eq:fcm-update} and figure~\ref{fig:DolphinFCM}(b) give $C(1) = \begin{pmatrix} 1 & 0 & 0 & 0 & 1 \end{pmatrix}$, $C(2) = \begin{pmatrix} 0 & 1 & 0 & 0 & 0 \end{pmatrix}$, $C(3) = \begin{pmatrix} 0 & 0 & 1 & 0 & 0 \end{pmatrix}$, and $C(4) = \begin{pmatrix} 0 & 0 & 0 & 1 & 0 \end{pmatrix} = C(0)$.
The state vectors $C(0)$, $C(1)$, $C(2)$, and $C(3)$ repeat themselves over and over in the same order. 
The dolphin FCM converges to a 4-state limit cycle. 
The active nodes in these states respectively are ``Survival Threat'', ``Herd Clustering'' and ``Run Away'', ``Fatigue'', and ``Rest''. 
So dolphins run away from threats like sharks in herd clusters. 
They then get tired and need to rest. 
The shark then catches up to the dolphins while they rest. 
This starts the cycle all over again. 

The set of all initial conditions $C(0)$ that lead to a given equilibrium describes the ``basin of attraction'' for that equilibrium. 
The FCM describes a map from these basins to their corresponding equilibrium attractors. 
The basins of the FCM's equilibria partition the FCM's input space. 
The FCM models a dynamical system by approximating its corresponding basin-to-equilibrium map. 

\subsection{Causal State Activation:  Clamping vs. Pulsing}

``Clamping'' the $k^{\text{th}}$ node $C_k$ to a value $c_k \in [0,1]$ forces $C_k(t)$ to equal $c_k$ irrespective of the values of other nodes at time $t-1$. 
The $k^{\text{th}}$ node is clamped ``on'' if $c_k \approx 1$ and it is clamped ``off'' if $c_k \approx 0$. 
Clamping acts as a forcing function and pushes the FCM into new equilibria. 
FCMs can clamp one or more nodes at a time. 
Clamping is often a way to implement policies and answer What-if questions regarding their effects. 

Clamping on the ``Survival Threat'' node changes the dolphin FCM's limit cycle from $\begin{pmatrix} 0 & 0 & 0 & 1 & 0 \end{pmatrix}$ $\rightarrow$  $\begin{pmatrix} 1 & 0 & 0 & 0 & 1 \end{pmatrix}$ $\rightarrow$ $\begin{pmatrix} 0 & 1 & 0 & 0 & 0 \end{pmatrix}$ $\rightarrow$ $\begin{pmatrix} 0 & 0 & 1 & 0 & 0 \end{pmatrix}$ $\rightarrow$ $\begin{pmatrix} 0 & 0 & 0 & 1 & 0 \end{pmatrix}$ to a new 3-state limit cycle $\begin{pmatrix} 0 & 1 & 0 & 1 & 0 \end{pmatrix}$ $\rightarrow$ $\begin{pmatrix} 1 & 0 & 0 & 1 & 0 \end{pmatrix}$ $\rightarrow$ $\begin{pmatrix} 1 & 1 & 0 & 1 & 0 \end{pmatrix}$ $\rightarrow$ $\begin{pmatrix} 0 & 1 & 0 & 1 & 0 \end{pmatrix}$. 
The active nodes in this new limit cycles respectively are ``Fatigue'' and ``Survival Threat''; ``Herd Clustering'' and ``Survival Threat''; and ``Herd Clustering'', ``Fatigue'', and ``Survival Threat''. 
The shark is \emph{always} present in this case. 
The dolphins cluster and get tired over and over. 
The nodes ``Rest'' and ``Run Away'' always off because the dolphins do not get a chance to rest and they never escape the shark. 

 FCMs can also `pulse' a node instead of clamping it. 
`Pulsing' the $k^{\text{th}}$ node $C_k$ to a value $c_k$ at time $t_0$ forces $C_k(t_0)$ to equal $c_k$ irrespective of the values of other nodes at time $t_0-1$. 
This only affects the FCM at time $t_0$ and the FCM reverts back to its unperturbed dynamics after $t_0$. 
The node $C_k$ pulses `on' at time $t_0$ if $C_k(t_0) \approx 1$ and it pulses `off' at time $t_0$ if $C_k(t_0) \approx 0$. 
FCMs can also pulse more than one node at a time. 

The unperturbed dolphin FCM may converge to its fixed point $\begin{pmatrix} 0 & 0 & 0 & 0 & 0\end{pmatrix}$. ``Survival Threat'' `on'-pulse can drive the FCM into a 4-step limit cycle instead. 

\subsection{FCM Sequencing}

FCM sequencing stimulates the FCM with a sequence of clamping patterns $\mathcal{A}_k$ to push it through a corresponding sequence of equilibria $\mathcal{B}_k$. 
This map from stimulus patterns $\mathcal{A}_k$ to their corresponding FCM equilibria $\mathcal{B}_k$ define a set of causal `if-then' meta-rules $\mathcal{A}_k\rightarrow\mathcal{B}_k$. 
The causal if-part $\mathcal{A}_k$ stimulates the FCM at the user's discretion through clamps and pulses. 
The arrow $\rightarrow$ describes the transient dynamics that connect $\mathcal{A}_k$ to its corresponding equilibrium $\mathcal{B}_k$. 
The stimulus pattern destroys the FCM's prior equilibrium and before pushing it into a new one. 

Figure~\ref{fig:figure4} shows this process for a sequence of 8 if-parts $\mathcal{A}_1$-$\mathcal{A}_8$ to the dolphin FCM. 
The figure also shows the corresponding then-parts
$\mathcal{B}_1$-$\mathcal{B}_8$. 
The transition from the if-part stimulus to the then-part equilibrium defines each of the 8 scenes in the virtual world video. 

Scene 1 clamps on ``Herd Clustering'' and shows the FCM converging to a 4-stage limit cycle. 
The dolphins play together as a group, get tired, rest, and then go back to playing. 
Scene 2 disrupts this equilibrium $\mathcal{B}_1$ by unclamping ``Herd Clustering and `on'-pulsing ``Survival Threat''. 
This moves the system from $\mathcal{B}_1$ to another 4-stage limit cycle $\mathcal{B}_2$ where the dolphins come across a shark, run away, get tired, rest, and encounter the shark again. 
Scene 3 then interrupts this equilibrium through stimulus $\mathcal{A}_3$. 
This process then repeats 5 more times with stimuli $\mathcal{A}_4$-$\mathcal{A}_8$. 
The FCM in turn converges to 2 more limit cycles $\mathcal{B}_4$ and $\mathcal{B}_8$ and 3 more fixed points $\mathcal{B}_5$-$\mathcal{B}_7$. 
The 8 scenes together tell the story of a pod of dolphins that gets chased by a persistent shark and has to push through fatigue to escape. 

\begin{figure}[ht]
\centering
\includegraphics[scale=0.38]{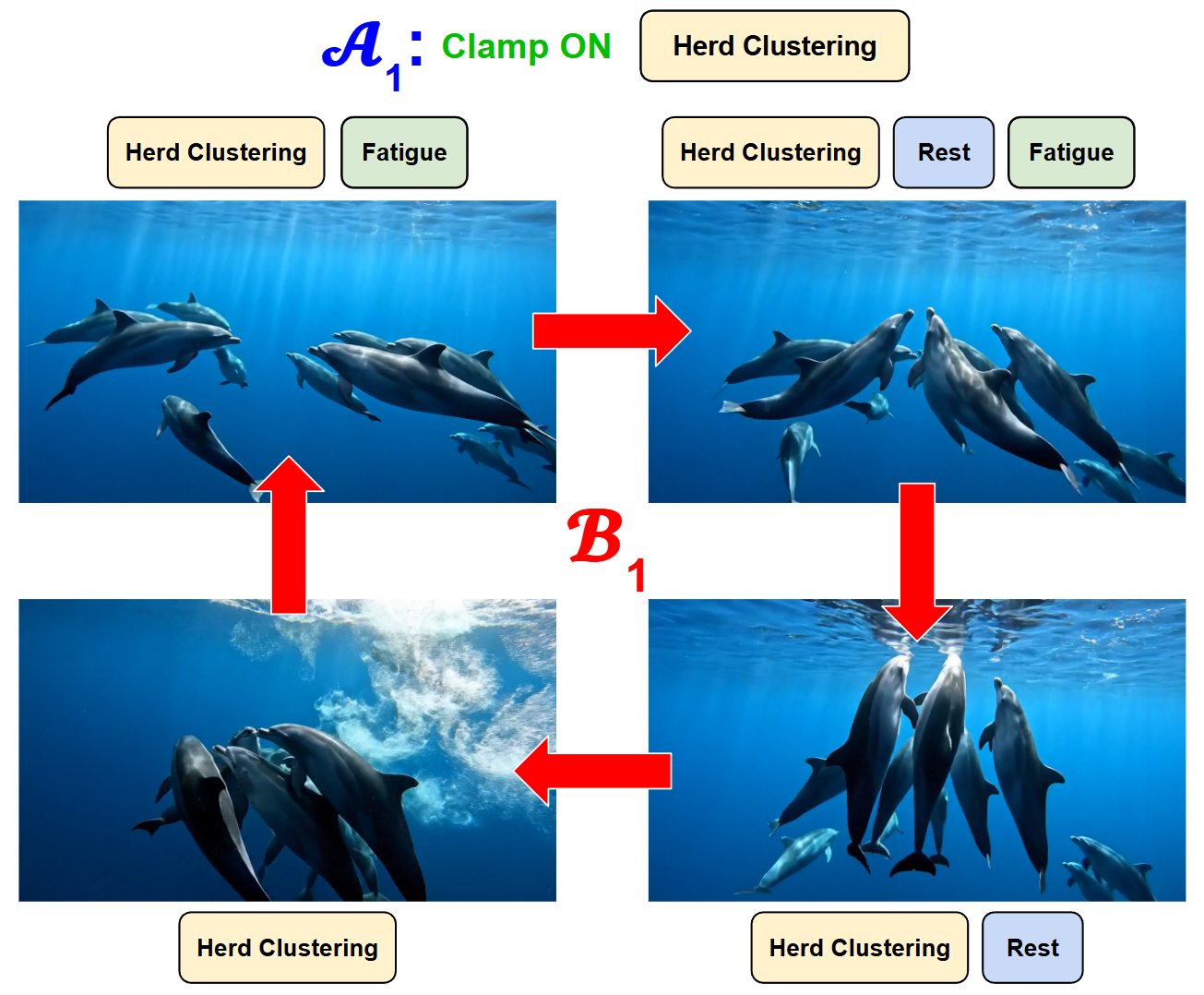}
\caption{Scene 1 from the virtual world video. 
$\mathcal{A}_1$ stimulates the dolphin FCM by clamping on the ``Herd Clustering'' node. 
The dolphin FCM the converges to a 4-stage limit cycle $\mathcal{B}_1$. 
The figure shows one frame from each stage of the limit cycle and also mentions their corresponding active nodes next to the frame. }
\label{fig:Scene1}
\end{figure}

\begin{figure}[ht]
\centering
\includegraphics[scale=0.36]{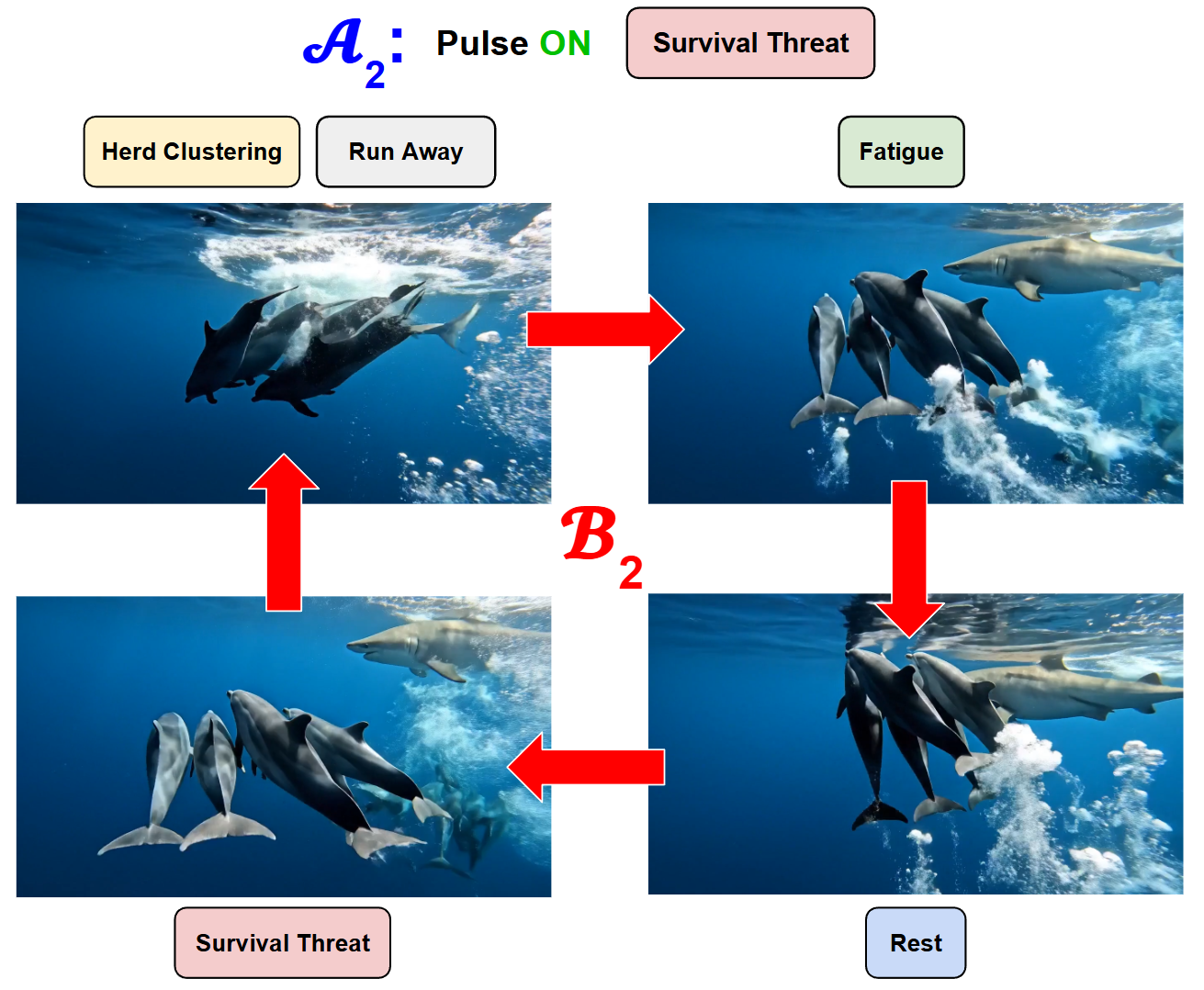}
\caption{Scene 2 from the virtual world video. 
$\mathcal{A}_2$ stimulates the dolphin FCM by pulsing on the ``Survival Threat'' node. 
The dolphin FCM the converges to a 4-stage limit cycle $\mathcal{B}_2$. 
The figure shows one frame from each stage of the limit cycle and also mentions their corresponding active nodes next to the frame. }
\label{fig:Scene2}
\end{figure}

\begin{figure}[ht]
\centering
\includegraphics[scale=0.36]{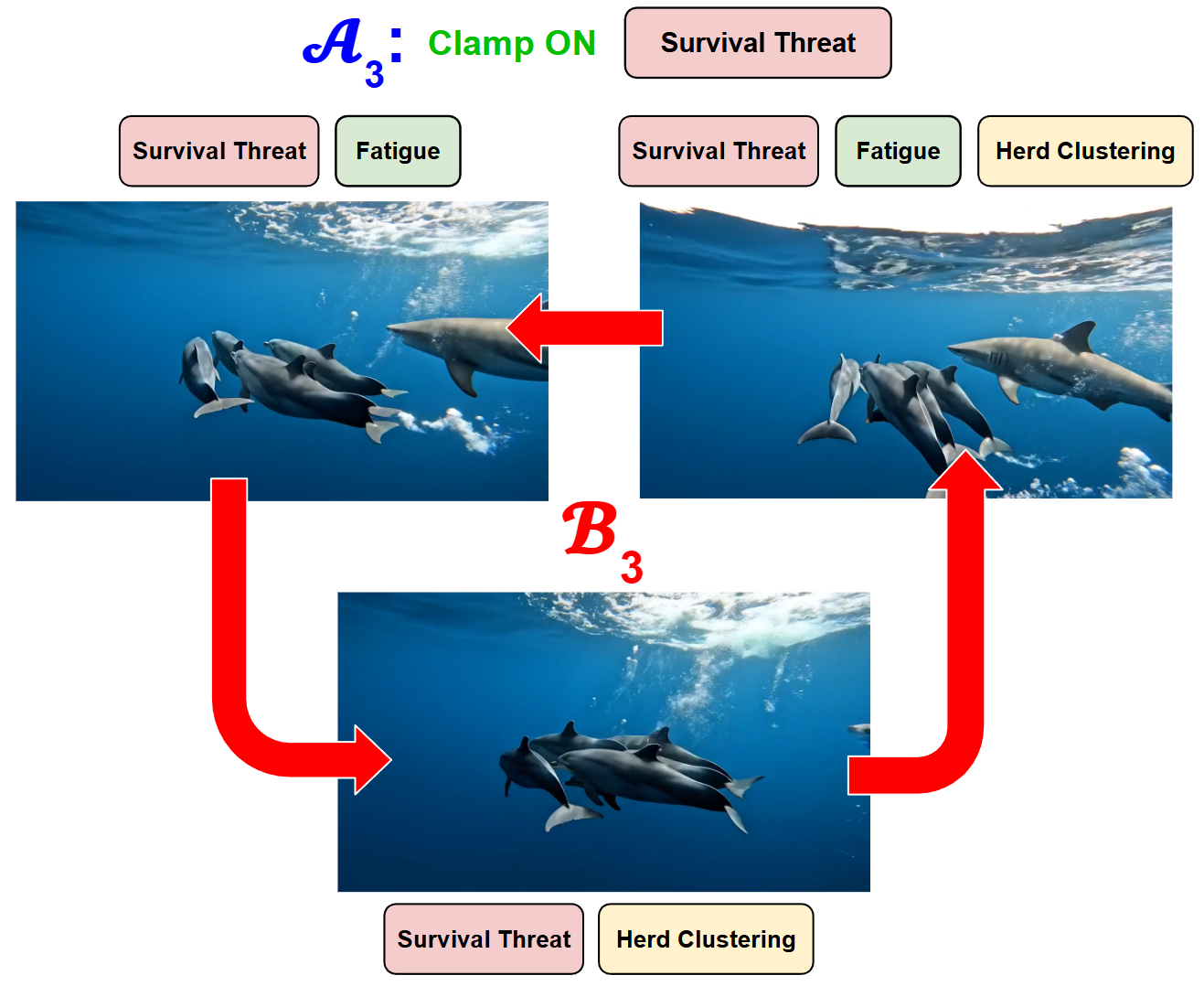}
\caption{Scene 3 from the virtual world video. 
$\mathcal{A}_3$ stimulates the dolphin FCM by clamping on the ``Survival Threat'' node. 
The dolphin FCM the converges to a 3-stage limit cycle $\mathcal{B}_3$. 
The figure shows one frame from each stage of the limit cycle and also mentions their corresponding active nodes next to the frame. }
\label{fig:Scene3}
\end{figure}

\begin{figure}[ht]
\centering
\includegraphics[scale=0.36]{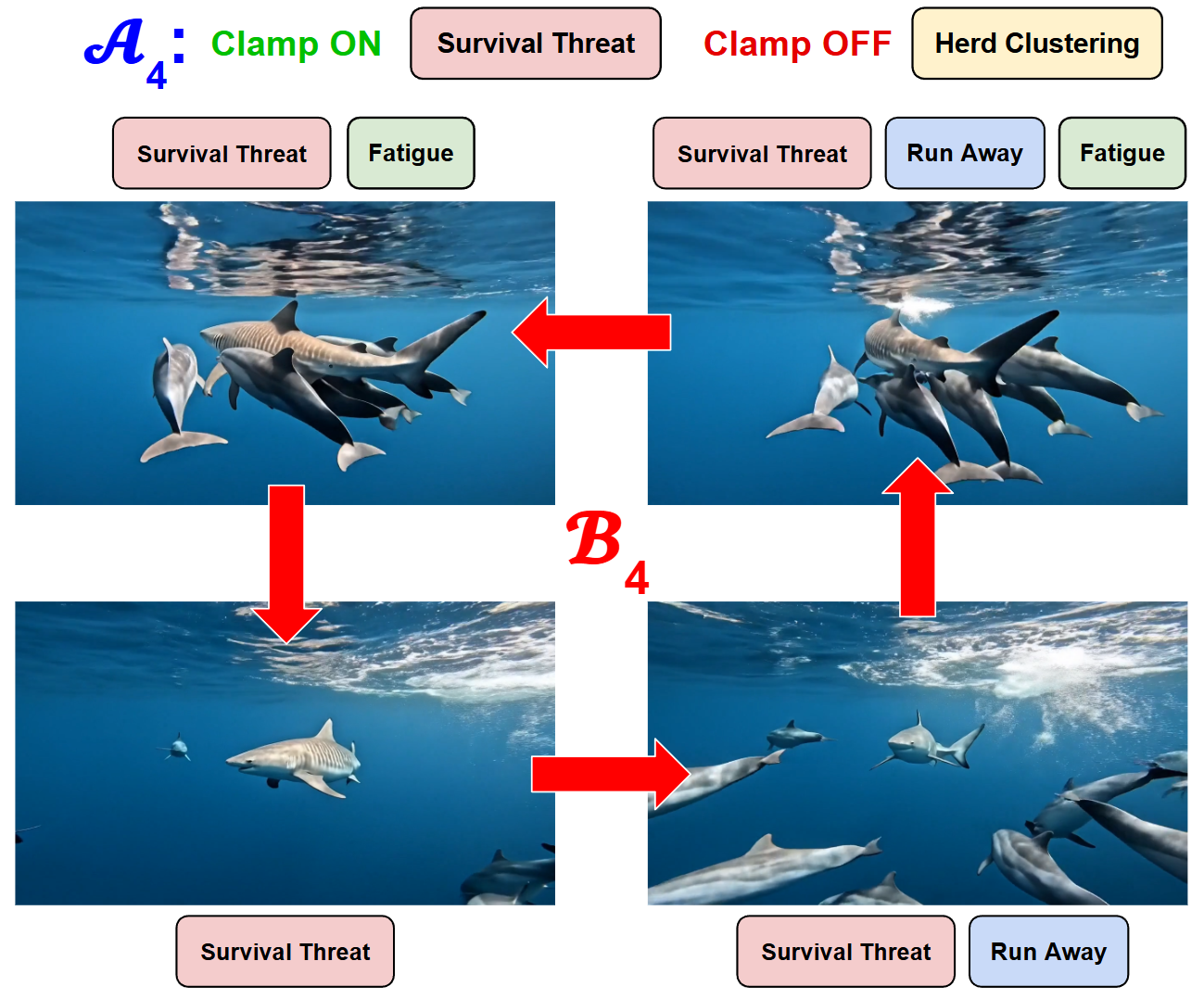}
\caption{Scene 4 from the virtual world video. 
$\mathcal{A}_4$ stimulates the dolphin FCM by clamping on the ``Survival Threat'' node as well as clamping off the ``Herd Clustering Node. 
The dolphin FCM then converges to a 4-stage limit cycle $\mathcal{B}_4$. 
The figure shows one frame from each stage of the limit cycle and also mentions their corresponding active nodes next to the frame. }
\label{fig:Scene4}
\vspace{-0.1in}
\end{figure}

Figures~\ref{fig:Scene1}-\ref{fig:Scene4} show the equilibria $\mathcal{B}_1$-$\mathcal{B}_4$ in detail. 
They show one frame from each state of the limit cycles. 
They also show the active nodes in those states. 
The limit cycle for $\mathcal{B}_8$ is the same as $\mathcal{B}_2$. 
The FCM equilibria $\mathcal{B}_5$-$\mathcal{B}_7$ are fixed points and thus have only one state. 
 Figure~\ref{fig:Figure2}(e)-(g) show the video frames that best represent those fixed-point states.

\begin{algorithm}[t]
\caption{Causal Virtual World Generation}
\label{alg:fcm_video}

\KwIn{$\mathcal{G}{=}(\mathcal{V}, \mathbf{E})$: FCM with $\mathbf{E}{\in}\mathbb{R}^{n \times n}$;
$\mathbf{c}^{(0)}{\in}\{0,1\}^n$: initial state and clamping conditions; $T$: time steps;
$\varphi(\cdot)$: threshold function;
$\mathcal{P}_{\text{script}}$: LLM scripting prompt;
$\mathcal{P}_{\text{video}}$: LVM generation prompt;
$\text{LLM}(\cdot)$, $\text{LVM}(\cdot)$: language and video models}
\KwOut{$\mathcal{D}$: causal virtual world documentary}

\BlankLine
\tcp{Step 1: FCM Simulation}
Initialize $\mathbf{C}{\leftarrow} \{ \  \} $, $\mathbf{c}{\leftarrow}\mathbf{c}^{(0)}$\;
\For{$t = 0$ \KwTo $T-1$}{
  Store $\mathbf{C}[\cdot,t] \leftarrow \mathbf{c}$, then update $\mathbf{c} \leftarrow \varphi(\mathbf{E} \cdot \mathbf{c})$\;
}
Detect equilibrium: set $\textit{eq} \leftarrow $ $\mathtt{FixedPoint}$ or $\mathtt{LimitCycle}$\;

\BlankLine
\tcp{Step 2: LLM Script Generation}
Analyze $\mathbf{C}$ for $\mathtt{START}$, $\mathtt{STOP}$, $\mathtt{PERSIST}$ transitions across all $t$\;
$\mathcal{SC} \leftarrow \text{LLM}(\mathcal{P}_{\text{script}},\, \mathcal{V},\, \mathbf{C},\, \textit{eq})$ 
\tcp*{yields $\mathtt{title}$ + $T$ $\mathtt{scene\ descriptors}$ $\{p_t, {lt}_t\}$}

\BlankLine
\tcp{Step 3: LVM Clip Generation and Assembly}
$\mathcal{D} \leftarrow \text{LVM}(\mathcal{P}_{\text{video}},\, \{p_t\},\, \{\textit{lt}_t\},\, \mathcal{SC}.\textit{title})$
\tcp*{ generate and concatenate clips}

\Return $\mathcal{D}$\;
\end{algorithm}

\section{Agentic Control of Large Video Generator Scene Prompts}

%OLIVER WRITES THIS SECTION.  Includes Vegas-like prompt instructors for the Gemini agent and full details of the simulations

We present the agentic control for generating the causal virtual world for an FCM.
This explains how to generate a video from a sequence of equilibria for an FCM with a large video generator. 
The sequence of equilibria comes from the concatenation of equilibria that result from the combination of input pulse signals and node clamping operation.
We tested our method with Google's Veo 3.1 video generator. 

Our method involves two main steps: screenwriting and movie production 
The screenwriting step converts an input sequence of equilibria into a movie script.
The script comprises multiple scenes. 
Each script can capture the causal information of a single state, a sequence of states, or a single equilibrium state.
The choice depends on multiple factors: the number of FCM nodes, the degree of information granularity in the visual world, the maximum video length of the video generator agent, and so on.

 An LLM agent with  screenwriting prompt $\mathcal{P}_{\text{script}}$ turns a list of active nodes in an FCM state vector into a sentence that describes the state based on the node names.
The agent turns the equilibrium state sequence into a paragraph that describes the FCM equilibrium.
The agent condenses that 1-paragraph description into a sentence that describes a scene. 
These scene descriptions form the script for the video generator.

The movie production feeds the movie script into the video generator agent and prompts it with $\mathcal{P}_{\text{video}}$ to create the video. 
This video generation can be a single continuous shot or a sequence of shots.
The single continuous shot generates all the scripts once. 
The maximum supported duration of the video  generator agent constrains this approach.
It may truncate important information.
The sequential approach decomposes the main script into multiple sub-scripts. 
The system first generates a video segment for the initial sub-script. 
Then it produces the second video segment conditioned on the first segment. 
Each subsequent generation conditions on the cumulative output of prior video segments. 
This recursive cycle continues until the sub-script sequence concludes.
The sequential method enables longer video durations than one-shot approaches.
Algorithm~\ref{alg:fcm_video} shows the complete method for script and video generation.

We present two agentic control methods based on autonomy levels. These include semi-autonomous and autonomous generations. 
Both methods utilize the two-step process of screenwriting and movie production.
But each method varies in autonomy level during the screenwriting step.

\subsection{Semi-autonomous agentic control}

An LLM agent executes the screenwriting phase of this agentic control method. 
This operator converts a sequence of equilibria into a script. 
Script scenes align with state transitions within the equilibria sequence. Each scene details granular information derived from active state variables or nodes. 
The agent reads the vector sequence directly or employs an intermediate visualization.
The agent inputs the script into the video generator agent. This agent transforms the script into a video or storyboard.

\begin{figure*}[!htb]
    \centering 
    \includegraphics[width=0.95\linewidth]{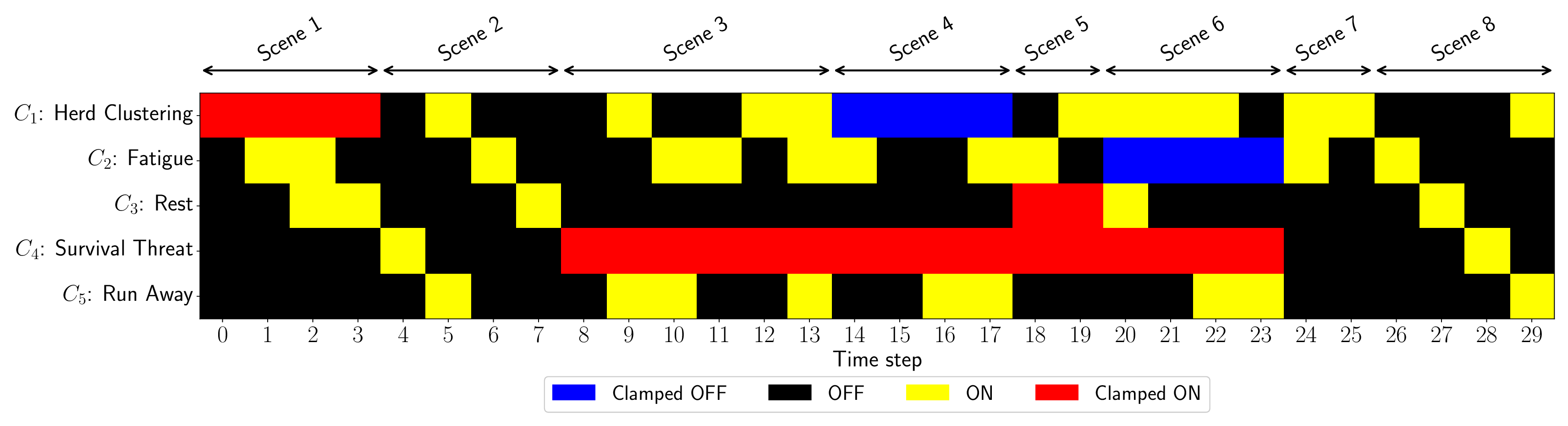} 
    \caption{Sequence of all FCM equilibria in the video:  Clamped-on states are in red and clamped-off states are in blue.} 
    \label{fig:seq_equilibria_visualization} % Adds a label for cross-referencing
    \vspace{-0.05in}
\end{figure*}

Let us consider the 5-node dolphin FCM in Figure \ref{fig:DolphinFCM}.  
Figure \ref{fig:figure4} shows  the links between a sequence of the dolphin FCM equilibria with pulsing and clamping. 
The sequence comprises of 8 equilibrium cases from the FCM.
Figure \ref{fig:seq_equilibria_visualization} visualizes the 8 equilibria and their transitions. 
Each scene corresponds to an equlibrium state.

An LLM agent annotated the sequenced equilibria into a movie script.
The movie script is divided into three parts: task description, summary, and the scene description for each equilibrium state.
The task description captures the high-level description of the movie. The summary provides more specific information. 
The scene descriptions translates the state variables into visual elements, dialogue and character interactions that advance the central narrative as laid out in the summary.

%Here is the task description and the summary for the sequenced equilibria in Figures \ref{fig:figure4} and \ref{fig:seq_equilibria_visualization}.

\begin{comment}
Here is the task description and summary from an annotation example for the sequenced equilibria in Figures \ref{fig:figure4} and \ref{fig:seq_equilibria_visualization}:
\begin{human_annotation}[title={Task description and summary},colbacktitle=gray!20, colback=green!10, colframe=black, boxrule=0.5pt,halign=justify,fonttitle=\fontsize{11}{13}\selectfont]
\noindent {\textbf{Task}}:  Create a 50 - 60 second video about a pod of 7 dolphins that encounter a large tiger  shark.  Use the following summary and the following sequence of 8 scene descriptions: \\

\noindent{{\textbf{Summary}}:   A pod of 7 dolphins rest and play in the ocean, encounter sharks, evade them, and then return to resting and playing. }
\end{human_annotation}
\end{comment}

Each scene in Figure \ref{fig:seq_equilibria_visualization} captures the causal transition from the causal if-part $\mathcal{A}_{k}$ to its corresponding equilibrium $\mathcal{B}_k$.
Consider the first pair ($\mathcal{A}_1$, $\mathcal{B}_1$).
The first column ${\bf C}(0) = \big{[}C_1(0), C_2(0), C_3(0), C_4(0), C_5(0)\big{]}$. 
The \emph{if-part} corresponds to clamping on the ``Herd Clustering" node while other nodes start in the OFF state.
The description of Scene 1 follow from the causal interaction from ${\bf C}(0)$ to ${\bf C}(3)$.

\begin{comment}
Here is an annotation example for scene 1:
\begin{human_annotation}[title={Scene 1 description from the annotation example}, colbacktitle=gray!20,colback=green!10, colframe=black, boxrule=0.5pt, halign=flush left,fonttitle=\fontsize{11}{13}\selectfont]
{{\textbf{Scene 1}}:  A pod of dolphins is resting and playing.}
\end{human_annotation}
\end{comment}

Here is a snippet of our prompt to the LLM for script generation.
\begin{human_annotation}[title={A snippet from the script-generation prompt $\mathcal{P}_{\text{script}}$}, colbacktitle=gray!20, colback=green!10, colframe=black, boxrule=0.5pt,halign=justify,fonttitle=\fontsize{11}{13}\selectfont]
\noindent{\textbf{Task Guidelines:}}\\
\\
\noindent{\textbf{1. Analyze the Data}: }\\
\noindent{\textbf{(a) Step $t=0$ (The Opening)}: Establish the scene based on which nodes are active (1) in the first column.} \\
\noindent{\textbf{(b) Transitions ($t=1$ to $t=T$)}: Compare column $t$ with column $t-1$.} \\
\noindent(i) Identify what {\textbf{Started} (0 to 1).} \\
\noindent(ii) Identify what {\textbf{Stopped} (1 to 0).} \\
\noindent(iii) Identify what {\textbf{Persisted} (1 to 1).} \\
\noindent{\textbf{(c) Interpret the causal logic}: If Node A is active at $t=1$ and Node B activates at $t=2$, imply that A might be causing B.}\\
\\
\noindent{\textbf{2. Documentary Structure}: }\\
\noindent{\textbf{(a) Title}: Generate a thematic title based on the Node List concepts.} \\
\noindent{\textbf{(b) Lower Thirds}: Provide short text overlays summarizing the state logic (e.g., "Rainfall triggers River Surge").} \\
\noindent{\textbf{(c) Narrative (Voiceover)}: A 1-2 sentence script connecting the current state to the previous one.}\\
\noindent{\textbf{(d) Conclusion}: Explicitly address the "Equilibrium Behavior" in the final scene.}
\end{human_annotation}

Google’s Veo 3.1 generative agent converted the LLM-generated script into a movie. 
The maximum allowed duration for a Veo 3.1 generated video is 8 seconds.
We used the multiple-shot approach to generate the video as a sequence of shots. 
Here is a snippet of the instruction we use for generating the video:
 \begin{human_annotation}[title={An snippet of the Veo 3.1 instruction $\mathcal{P}_{\text{video}}$},colbacktitle=gray!20, colback=pink!10, colframe=black, boxrule=0.5pt, halign=justify,fonttitle=\fontsize{11}{13}\selectfont]
\noindent{{\textbf{Style:}}  
Cinematic 4K, underwater photography, realistic Pacific Bottlenose dolphins, deep ocean blue color grade, naturalistic lighting, fluid camera movements, high-fidelity water physics. \\

\noindent{\textbf{Prompt:}} Generate A pod of dolphins is seen from a side-angle tracking shot, swimming at maximum speed through the deep blue ocean. Intense cavitation bubbles trail from their fins. The camera moves rapidly alongside them. The scene ends with the dolphins beginning to slow down. \\

\noindent{\textbf{Transition Note:}} End shot with decelerating tail beats.}
\end{human_annotation}
We sequentially prompted the video generator with the scene descriptions from the LLM annotated movie script.
The agent rendered the Scene 1 description as an 8-second clip. 
The model extended the output to 16 seconds using the Scene 2 description. 
This recursive extension continued through all 8 scene descriptions. 
The procedure yielded a 57-second video encapsulating all scenes.
\begin{comment}
Here is the prompt for generating the video for Scene 1:
\begin{llmbox}[title={Prompt: Generate the video Scene 1 with Veo 3.1}, colbacktitle=gray!20,colback=yellow!10, colframe=black, boxrule=0.5pt, halign=flush left, fonttitle=\fontsize{11}{13}\selectfont]
\centering{{\textbf{Scene 1}}:  A pod of dolphins is resting and playing.}
\end{llmbox}
\end{comment}

Figure ~\ref{fig:Figure2} shows frames from the generated video.
Figure ~\ref{fig:shark_dolphin_scene_1} shows a frame from Scene 1 and corresponds to its equilibrium state in Figure \ref{fig:seq_equilibria_visualization}.
This scene runs from $t=0$ to $t=3$ with ``Herd Clustering'' node clamped ON and the ``Survival Threat'' node inactive:
dolphins play and rest without a survival threat.
Here is the annotation for this scene: \emph{A pod of dolphin is resting and playing}.
Figure ~\ref{fig:shark_dolphin_scene_1}  depicts this.
Figure ~\ref{fig:shark_dolphin_scene_2} - \ref{fig:shark_dolphin_scene_8} similarly capture the corresponding causal virtual world of Scene 2-8 in Figure~\ref{fig:seq_equilibria_visualization}.

Again we can use LLM agents to extract FCMs from text or vice versa \cite{panda2025causal, agenticleash2025} and exploit the Bayesian \emph{de-chunking} structure \cite{panda2026agentic}.
This allows AI agents to partially or fully automate the selection of scenes or meta-rules $\mathcal{A}_j \rightarrow \mathcal{B}_j$.

%We used Google's Veo 3.1 generative agent to transform the movie script into a movie. 
%The agent used the multiple shot approach to crate the video.
%It transformed Scene 1 description into an 8 second video.
%It then extend the movie to a 16 seconds video  by using the Scene 2 description. 
%It further extended that with the Scene 3 description.
%We ran this recursive video extension runs until we have gone through the 8 scene decriptions.
%This process produced a 57 second video that captures the  all the scenes.

\section{Conclusions}\label{conclusion}
Sequenced FCMs can combine with agentic control of large video generators to produce causal virtual worlds or videos.
Users can define or modify the FCMs or oversee LLMs that generate them from text or from transcribed speech.
Users can also modify the virtual world midstream if they impose new causal patterns or if they adapt the FCM's local causal structure.
These techniques scale and favor hardware implementation both because FCM causal propagation involves only simple vector-matrix multiplication of often sparse causal edge matrices and because mixing FCMs always produces a new FCM.
Large-scale versions offer in principle a way to design realistic immersive virtual worlds that allow multiple users to interact.

\bibliographystyle{unsrt}
\bibliography{bibdata}

\end{document}